\documentclass{article}

\usepackage{arxiv}

\usepackage[utf8]{inputenc} 
\usepackage[T1]{fontenc}    
\usepackage{hyperref}       
\usepackage{url}            
\usepackage{booktabs}       
\usepackage{amsfonts}       
\usepackage{nicefrac}       
\usepackage{microtype}      
\usepackage{lipsum}
\usepackage{graphicx}
\usepackage{cite}
\usepackage{amsmath,amssymb,amsfonts}
\usepackage{algorithmic}
\usepackage{graphicx}
\usepackage{textcomp}
\usepackage{tikz}
\usepackage{relsize,exscale}
\DeclareMathOperator{\atantwo}{atan2}

\title{A comparative study of human inverse kinematics techniques for lower limbs}

\author{
 Zineb BENHMIDOUCH, Saad MOUFID, Aissam AIT OMAR \\
\texttt{z.benhmidouch@gmail.com}}

\begin{document}
\maketitle

\begin{abstract}
Inverse Kinematics (IK) remains a dynamic field of research, with various methods striving for speed and precision. Despite advancements, many IK techniques face significant challenges, including high computational demands and the risk of generating unrealistic joint configurations. This paper conducts a comprehensive comparative analysis of leading IK methods applied to the human leg, aiming to identify the most effective approach. We evaluate each method based on computational efficiency and its ability to produce realistic postures, while adhering to the natural range of motion and comfort zones of the joints. The findings provide insights into optimizing IK solutions for practical applications in biomechanics and animation.
\end{abstract}
\section{Introduction}
\label{sec:introduction}

One of the most critical and challenging aspects in developing robots aimed at restoring human mobility after neurological injuries is solving the Inverse Kinematics (IK) of physiological limbs. This process involves calculating joint angles based on predefined workspace coordinates. The complexity of the IK problem is influenced by the manipulator’s geometry and the nonlinearity of its model, which defines the relationship between task and joint spaces. For real-time control, precise IK solutions are essential for the robot to perform tasks effectively. IK techniques are generally classified into three categories: analytical methods, numerical methods, and intelligent methods.
\\
The analytical method addresses Inverse Kinematics (IK) by solving a set of closed-form equations, which directly compute the generalized coordinates needed to position the manipulator’s end effector at a predefined target location~\cite{b1}. This method leverages geometric insights and the specific structure of the robot. However, for arbitrary robotic kinematics, analytical solutions may either be non-existent or multiple. In contrast, numerical methods are iterative approaches that converge to a single solution based on an initial estimate. Common algorithms in numerical IK include the Moore-Penrose pseudo-inverse method~\cite{b2}, cyclic coordinate descent method~\cite{b3}, Levenberg-Marquardt damped least squares method~\cite{b4}, optimization methods, and multi-objective optimization using genetic algorithms~\cite{b6}. Additionally, the neural network approach~\cite{b7} explores the entire configuration workspace of the robot to determine the optimal solution.
\\
The primary advantage of the analytical method is its accuracy and efficiency, providing real-time results and computing valid potential configurations. However, as the number of degrees of freedom in a manipulator increases beyond six, the number of possible solutions becomes very large. Furthermore, if a solution is not feasible, the analytical method does not provide an approximate solution that meets all constraints. In contrast, methods like the pseudo-inverse and damped least squares often suffer from poor performance and slow computation times. While the cyclic coordinate descent method is relatively simple and computationally fast, it struggles with generating smooth motions and incorporating non-geometric constraints such as minimum energy criteria. Optimization-based methods address IK problems in a discrete, point-to-point manner, unlike the neural network approach, which offers a more holistic solution by exploring the entire configuration workspace.
\\
This paper investigates various Inverse Kinematics (IK) algorithms tailored for lower limb applications to determine the most effective method based on efficiency, accuracy, computational cost, energy consumption, and the ability to produce realistic postures. We evaluate analytical, numerical, and intelligent methods to solve the generalized IK problem, considering position, orientation, and angular constraints. By examining these approaches, we aim to identify the optimal technique for balancing performance and practicality in real-world applications. 
\\
The structure of the paper is organized as follows: Section \ref{Articulated human leg model} outlines the mechanical structure of the human lower limb, including an in-depth discussion of forward kinematics, workspace analysis, and trajectory planning. Section \ref{Inverse kinematics of the human leg} explores the application of various inverse kinematics methods to the human leg. Finally, Sections \ref{Simulation results} and \ref{Discussion and Conclusion} provide and analyze the simulation results, leading to the concluding discussion.
\section{ARTICULATED HUMAN LEG MODEL}
\label{Articulated human leg model}
\begin{figure}
    \centering
    \includegraphics[width=0.7\textwidth]{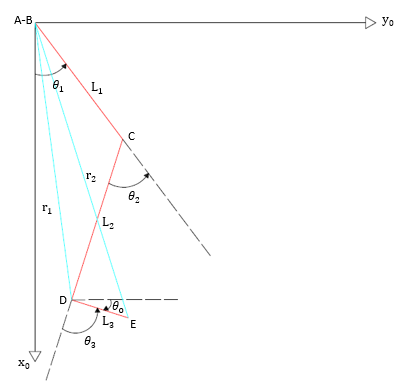}
    \caption{Model of the right lower limb in a sagittal plane}
    \label{fig:1}
   \end{figure}

\begin{figure}
    \centering
    \includegraphics[width=0.7\textwidth]{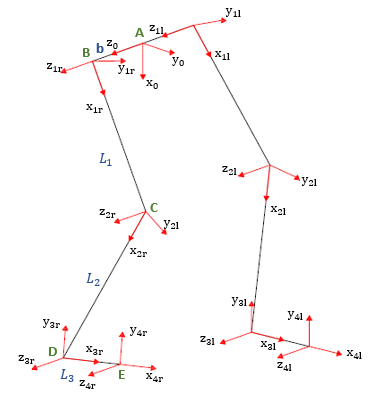}
    \caption{Physiological lower limbs diagram}
    \label{fig:2}
   \end{figure}

This section outlines the functional description of the lower limbs based on Calais Germain's work~\cite{b9}. In biomechanics, movement is analyzed using a coordinate system with three planes: the sagittal plane (median plane), the frontal plane, and the transverse plane. These planes intersect along three axes: the left-right axis (frontal axis), the anteroposterior axis (sagittal axis), and the craniocaudal axis (vertical axis).\\
Anatomically, the lower limb is divided into four segments: the pelvis, thigh, leg, and foot, connected by three major joints: the hip (coxofemoral joint), the knee, and the ankle. The hip joint, which functions as a spherical joint, allows rotation around all three axes, providing three degrees of freedom (DOF). The knee joint offers 1-DOF, permitting flexion and extension in the sagittal plane around the left-right axis. The ankle joint enables movement in all three planes, resulting in 3-DOF. Therefore, each lower limb has a total of 7-DOF. For the purposes of this study, we focus exclusively on movements within the sagittal plane, specifically flexion and extension, which involves 3-DOF per lower limb.
\\
The kinematics of the physiological lower limbs is used to establish a relationship between the Cartesian coordinates of the big toe $E$ and the generalized coordinate $q=\left[\begin{array}{ccc} \theta_1 & \theta_2 & \theta_3 \end{array}\right]^T $. To achieve this, we utilize the modified Denavit-Hartenberg (D-H) convention to describe the reference frame at each joint of the lower limbs, as illustrated in Figures \ref{fig:1} and \ref{fig:2}. The link parameters for the right kinematic chain are detailed in Table \ref{tab:1}. Consequently, the relationship between the coordinate system of the big toe $E$ and the joint angular displacements is represented by the following matrix.
\begin{table}
\centering
\caption{Modified D-H parameters for the right lower limb}
 \label{tab:1}
\begin{tabular}{|c|c|c|c|c|}
\hline
        Joint & $\alpha_{i-1}$ & $a_{i-1}$ & $d_i$ & $\theta_i$  \\
          \hline
         1 & 0 & 0 & b & $\theta_1$  \\
        
         2 & 0 & $L_1$ & 0 & $-\theta_2$ \\
          
         3 & 0 & $L_2$ & 0 & $\theta_3$\\
           
         4 & 0 & $L_3$ & 0 & 0 \\
         \hline
    \end{tabular}
\end{table}
   
\begin{equation}
T_4^{0}=\left[\begin{array}{cccc}
    
     C_{1-23} &   -S_{1-23}  & 0 &  L_3.C_{1-23} +L_2.C_{1-2} +L_1.C_{1}\\
     S_{1-23} &   C_{1-23}  & 0 & L_3.S_{1-23}+L_2.S_{1-2}+L_1.S_{1}\\
      0 & 0 & 1 & b\\
      0 & 0 & 0 & 1
\end{array}\right] 
\label{eq:100}
\end{equation}
\\Where,\\
$C_{1-23} :=\cos(\theta_{1}-\theta_{2}+\theta_{3});$
\\$S_{1-23} :=\sin(\theta_{1}-\theta_{2}+\theta_{3});$
\\$C_{1-2} :=\cos(\theta_{1}-\theta_{2});$
\\$S_{1-2} :=\sin(\theta_{1}-\theta_{2});$
\\$C_{1} :=\cos(\theta_{1});$
\\$S_{1} :=\sin(\theta_{1});$\\
The Monte Carlo method \cite{b12} is employed to determine the feasible workspace of the lower limbs and assess their operational capabilities. The method operates on the following principle: it generates pseudo-random numbers uniformly distributed within the interval $[0,1]$ Each random sample then provides a set of variable values for the generalized coordinate $q$ using the equation:
\begin{equation} \label{Eq:2}
     q = q_{min} + \rho(q_{max} -q_{min} ) 
\end{equation}
Where: 
\\$ q_{min}$, $ q_{max}$ : Minimum and maximum range of  motion;
\\ $\rho$ : pseudo-random number in the interval $[0,1]$ which respects a uniform distribution;
\\$q $ : the generalized coordinate random value obtained by Monte Carlo method;\\
This approach allows us to produce a graphical representation of the lower limb's workspace in the sagittal plane, while adhering to the joint constraints specified in Table \ref{tab:2}, as illustrated in Figure \ref{fig:11}.
\begin{table}
    \centering
    \caption{Range of motion of the lower limb joints \cite{b13}}
    \begin{tabular}{|c|c|c|c|c|}
      \hline
        Joint & $\theta_i^{min}$ & $\theta_i^{max}$ & Comfort zone &  Conditions \\
          \hline
         i=1 & -20° & 120° & from 15.75° to 39.55° & knee neutral 0°  \\
         i=2 & 0° & 118° & from 0° to 39.55° & hip neutral 0° \\
         i=3 & 50° & 126° & from 77.75° to 103.3° & knee neutral 0°\\
         \hline
    \end{tabular}
    
    \label{tab:2}
\end{table}

Additionally, the comfort zone represents a subset of the joint's total range of motion, defined as $35\%$ of this range. The center of the comfort zone $q^{comf}_c$,  is calculated using the following equation \cite{b13}:
\begin{equation}\label{eq:3}
    q^{comf}_c=\frac{q^{comf}_{max}-q^{comf}_{min}}{2}+q^{comf}_h
\end{equation}
where $q^{comf}_{max}$ and $q^{comf}_{min}$ denote the maximum and minimum values of the comfort zone, and $q^{comf}_h$ 
  represents the home position of the generalized coordinate $q$.
\\
Additionally, trajectory planning is proposed using the minimum jerk criterion. In this context, jerk is defined as the third derivative of position with respect to time. Applying the jerk criterion helps ensure continuous acceleration of the joints, which reduces vibrations and avoids resonance frequencies \cite{b14}. The minimum jerk criterion is achieved by minimizing the following function:
\begin{equation}
J = \frac{1}{2}\int_{0}^{T}{\left(\frac{d^3x(t)}{dt^3}\right)^2}dt
\label{eq:4}
\end{equation}
To satisfy the minimum jerk criterion, the position $x(t)$ is described by a fifth-order polynomial, expressed as follows:
\begin{equation}\label{eq:5}
    x(t)= s_0+s_1 t+s_2 t^2+s_3 t^3+s_4 t^4+s_5 t^5
\end{equation}
This polynomial formulation allows for the specification of position, velocity, and acceleration at both the start and end of the motion. The final polynomial coefficients are determined by solving the boundary conditions, as detailed in Equation (\ref{eq:6}) \cite{b27}.
    
\begin{equation}\label{eq:6}
    \left[\begin{array}{c}
    
     s_0 \\
     s_1  \\
     s_2\\
      s_3\\
      s_4\\
      s_5\\
\end{array}\right]=
\left[\begin{array}{cccccc}
    
     1 & t_0 & t_0^2 & t_0^3 & t_0^4 & t_0^5 \\
     0 & 1 & 2.t_0 & 3.t_0^2 & 4.t_0^3 & 5.t_0^4  \\
     0 & 0 & 2 & 6.t_0 & 12.t_0^2 & 20.t_0^3\\
      1 & t_f & t_f^2 & t_f^3 & t_f^4 & t_f^5\\
      0 & 1 & 2.t_f & 3.t_f^2 & 4.t_f^3 & 5.t_f^4 \\
      0 & 0 & 2 & 6.t_f & 12.t_f^2 & 20.t_f^3\\
\end{array}\right]^{-1}
\left[\begin{array}{c}
    
     x_0 \\
     \dot{x}_0  \\
     \ddot{x}_0 \\
      x_f \\
     \dot{x}_f  \\
     \ddot{x}_f  \\
\end{array}\right]
\end{equation}
\section{Inverse kinematics of the human leg}
\label{Inverse kinematics of the human leg}

In \cite{b16}, generating a trajectory in joint space is shown to facilitate effective control of the lower limbs while circumventing issues associated with kinematic singularities. This approach also addresses challenges related to manipulator redundancy, as seen in the proposed human leg model. Redundancy enables the lower limbs to bypass kinematic constraints and reduce energy consumption. Consequently, inverse kinematics can be used to derive the corresponding trajectory in joint space after it has been defined in task space. In our case, the reference system $R_0$ is treated as stationary relative to an inertial reference frame $R_U$ to avoid an infinite number of solutions. This section explores various methods for solving the inverse kinematics problem for the human leg.
\subsection{Analytical method}
Once the forward kinematics are defined, the next step is to determine the inverse kinematics (IK). This involves finding the joint angles \(\theta_1\), \(\theta_2\), and \(\theta_3\) given the predefined coordinates of the end effector \(E (E_x, E_y)\) and its orientation relative to the transverse plane, \(\theta_0\). As illustrated in Figure \ref{fig:2}, the coordinate space of the ankle joint \(D\) is given as follows:
\\
\begin{equation}
\label{eq:8}
    \bigg\{\begin{array}{c}
     D_x=E_x-L_3.S_0 \\
     D_y=E_y-L_3.C_0
\end{array}
\end{equation}
\\
Using D-H modified method, the coordinate space of ankle joint is given by:
\begin{equation}
\label{eq:9}
    \bigg\{\begin{array}{c}
     D_x=L_2.C_{1-2}+L_1.C_{1} \\
     D_y=L_2.S_{1-2}+L_1.S_{1}
\end{array}
\end{equation}
Equations (\ref{eq:8}) and (\ref{eq:9}), yields to:
\begin{equation}\label{eq:10}
    C_2=\frac{(E_x-L_3.S_0)^2+(E_y-L_3.C_0)^2-L_1^2-L_2^2}{2.L_1.L_2}
\end{equation}
\begin{equation}\label{eq:11}
    S_2=\pm\sqrt{1-{C_2}^2}
\end{equation}
Then,\\
\begin{equation}\label{eq:12}
    \theta_2=\atantwo (S_2,C_2)
\end{equation}
By using this trigonometric equation, we find:
\begin{equation}
\label{eq:14}
    \theta_1= \atantwo(E_x-L_3.S_0,E_y-L_3.C_0)-\atantwo(L_2.S_2,L_1+L_2.C_2)
\end{equation}
and 
\begin{equation}\label{eq:15}
    \theta_3= \frac{\pi}{2}-\theta_1+\theta_2-\theta_o
\end{equation}
However, if the orientation angle of the foot is not known, there are multiple possible solutions.

\subsection{Numerical methods}
\subsubsection{Cyclic coordinate descent method}
Cyclic Coordinate Descent (CCD) is an iterative algorithm used to solve inverse kinematics problems. Yotchon et al. \cite{b24} proposed a hybrid approach combining the CCD method with a differential evolution algorithm, a metaheuristic optimization technique, to tackle inverse kinematics challenges. This combined method reliably converges to the target position regardless of the initial conditions. The CCD algorithm works by minimizing joint errors through iterative adjustments of one component of the angular vector at a time. While it can be used in real-time applications, it typically requires multiple iterations to achieve the desired outcome \cite{b17}. One of the key benefits of CCD is its simplicity in implementation. The algorithm involves measuring the discrepancy between the target position and the end-effector position, then applying a rotation matrix to reduce this discrepancy to zero, all while considering joint constraints. This process is repeated sequentially for each joint, starting from the end-effector and moving towards the root joint of the kinematic chain.   
\subsubsection{Moore-Penrose pseudo-inverse method}
The Moore-Penrose Pseudo-Inverse method (MPPI) is based on the Newton-Raphson approach and involves solving the following nonlinear equation:
\begin{equation}
    N(q_d)\equiv f(q_d)-p_d=0
\end{equation}
where $f$ represents the forward kinematics equation, $p_d$ is the target position and orientation, and $q$ is the generalised coordinates. Klein et al. \cite{b25} demonstrated that the pseudo-inverse IK method is not inherently repeatable. This implies that the algorithm can be computationally inefficient and does not guarantee a minimum norm for angular joint velocities.
\\
To generalize the Newton-Raphson procedure for solving inverse kinematics, we utilize the Taylor expansion of the forward kinematics function around $q_d$. The expansion is given by:
\begin{equation}
    p_d= f(q_d)=f(q_k)+ \frac{\partial{f}}{\partial{q}}\bigg|_{q_k}(q_d-q_k)
\end{equation}
Thus
\begin{equation}\label{eq:50}
    \Delta q= J^+(q_k)(p_d-f(q_k))
\end{equation}
Where, 
\begin{equation}
    J=\frac{\partial{f}}{\partial{q}}
\end{equation}
The pseudo-inverse of the lower limb's Jacobian matrix \( J \in \mathbb{R}^{2 \times 3} \), denoted as \( J^+ \), is given by:
\begin{equation}
    J^+=J^T (JJ^T)^{-1}
\end{equation}
based on (\ref{eq:50}), an iterative method can be used to calculate the joints update :
\begin{equation}
    q_{k+1}=q_k+J^+(q_k)(p_d-f(q_k))
\end{equation}
Where $q_k$ and $f(q_k)$ are the generalized coordinates and the end effector position at iteration k, respectively. 
\subsubsection{Levenberg Marquardt damped least squares method}

A method to address the pseudo-inverse problem is the Levenberg-Marquardt Damped Least Squares (LMDLS) method. Wampler et al. \cite{b26} proposed an approach to determine the optimal damping factor, which balances the angular joint velocities with the tracking error. This approach involves finding the joint angular error vector $\Delta q$ that minimizes the tracking error and the joint velocities. This is achieved by minimizing the following objective function:
\begin{equation}\label{eq:51}
    \lVert \Delta x-J\Delta q\lVert ^2-\lambda^2 \lVert\Delta q\lVert ^2
\end{equation}
Where $\lambda >0$ is the damping factor.
\\
Solving (\ref{eq:51}), the damped pseudo-inverse $J^{+\lambda}$ is as follows: 
\begin{equation}
    J^{+\lambda}=J^T(J^TJ+\lambda I_{(3\times3)})^{-1}
\end{equation}
\\
Thus the joints' updates are giving by the following equation:
\begin{equation}
    q_{i+1}=q_i+J^{+\lambda}(q_i)(p_d-f(q_i))
\end{equation}
Several methods have been proposed in the literature to select an optimal damping factor $\lambda$ for the Levenberg-Marquardt Damped Least Squares (LMDLS) method, making it more robust against singularities and solvability issues \cite{b18}. Among these methods, one approach is to use a constant value for$\lambda$ \cite{b19}. However, to account for the natural limits of human posture and ensure that joint angles stay within the bounds, $q_{min}<q<q_{max}$, the diagonal matrix $\lambda I_{(3\times3)}$ is replaced by $D(\lambda)$, \cite{b10}.
\begin{equation}
    D(\lambda)=\left[\begin{array}{cccc}
    
     \lambda_1 &   0 & 0 \\
     0 &   \lambda_2 & 0\\
     0 &   0 & \lambda_3 \\
     
\end{array}\right] 
\end{equation}
Where $\lambda =\left[ \begin{array}{ccc} \lambda_1 & \lambda_2 & \lambda_3 \end{array}\right]^T$,  is defined as follows:
\begin{equation}
    \lambda=a{\left(\frac{2(q-q^{comf}_c)}{q_{max}-q_{min}} \right)}^b
\end{equation}  
where \(a\) and \(b\) are positive constants, and \(q^{comf}_c\) is the center of the comfort zone for joint configurations \(q\). The limits of the comfort zone are calculated as \(0.35\) times the range of the joint's total motion. Specifically: 
\begin{equation}
    \text{Comfort Zone Limits} = 0.35 \times (\text{Joint Range of Motion})
\end{equation}
\\
Thus, $\lambda$  imposes a constraint on each joint angular value $\theta_i$, $i\in \left\{1,2,3\right\}$. When $\theta_i$ is within its specified range of motion, as detailed in Table \ref{tab:2}, a small value of $\lambda_i$  produces accurate results. Conversely, if $\theta_i$ approaches its limits, a larger value of $\lambda_i$ helps ensure a feasible solution.
\subsubsection{Optimization method}
\label{Optimization method}
Tringali et al. \cite{b22} introduced an optimal inverse kinematics approach based on optimization techniques for redundant robot manipulators, incorporating both linear and nonlinear constraints by selecting appropriate initial conditions. Similarly, Lu \cite{b23} employed optimization methods to ensure feasible and smooth joint motion while also addressing collision detection in the workspace.
\\
Constrained optimization is based on finding an angular vector $q=\left[\begin{array}{ccc} \theta_1 & \theta_2 & \theta_3 \end{array}\right]^T $, that is a local minimum to the objective function $f(q)$ given as follows:
\begin{equation}
    \begin{array}{c}
     f(q)=|\sqrt{x_d^2+y_d^2}-r_2| \\
     \\r_2=\sqrt{x^2+y^2}\\
     \\x=L_3C_{1-2+3}+L_2C_{1-2}+L_1C_{1}\\
     \\y=L_3S_{1-2+3}+L_2S_{1-2}+L_1S_{1}\\  
\end{array}
\end{equation}
Where $x_d$ and $y_d$ represent the target position coordinates.
The primary objective is to determine a feasible configuration while adhering to various constraints, including the restriction of joint angular limits. This is achieved using a barrier function to enforce these constraints.
\begin{equation}   
B_k(q)=f(q)+\frac{1}{k}b(q)
\end{equation}
This method is known as the interior point algorithm. As the angular vector \( q \) approaches the boundary of the feasible region, the barrier function \( b(q) \) tends towards \( +\infty \). The logarithmic barrier function \( b(q) \) is given by:
\begin{equation}\label{log}   
b(q)=-(\log(-(q-q_{max}))+\log(-(q_{min}-q)))
\end{equation}
The other constraints can be expressed as a distance as follows:
\begin{equation}
    \bigg\{\begin{array}{c}
    x_d=x \\
     \\y_d=y\\
\end{array}
\end{equation}
\subsubsection{ Multi-objective optimization genetic algorithm method}

A multi-objective optimization genetic algorithm (MOOGA) is a metaheuristic technique inspired by evolutionary biology processes, designed to tackle optimization problems. Bjoerlykhaug \cite{b22} utilized MOOGA to solve inverse kinematics in real-time while the robot is in motion, resulting in a reduction of computational time by $50\%$. Solving an optimization problem using a genetic algorithm involves three primary tasks:
\begin{enumerate}
    \item Initial Random Population: Generating an initial set of candidate solutions.
    \item Genetic Operators: Applying operators such as selection, crossover, and mutation to evolve the population.
    \item Objective Evaluation Function: Assessing the quality of each candidate solution according to the defined objectives.
\end{enumerate}
This approach allows for efficient and effective exploration of the solution space, especially in complex, multi-objective scenarios.
\\
\subsection{Neural networks method}
\begin{figure}
    \centering
    \includegraphics[width=0.7\textwidth]{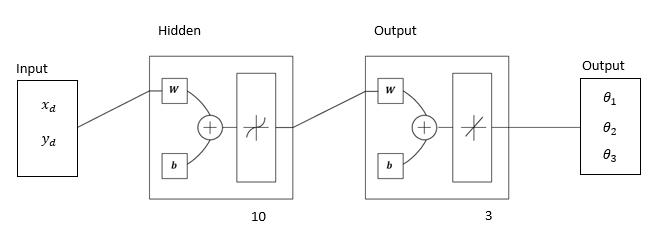}
        \caption{Neural network structure}
    \label{fig:27}
\end{figure}

This subsection explores the use of neural networks to find inverse kinematics (IK) solutions for the human leg. Shah et al. \cite{b20} applied deep artificial neural networks to solve the IK problem for a 5-axis serial link robotic manipulator. They achieved an accuracy where the deviation between the end effector and the target, due to actuator limitations, was approximately 0.2 mm; this deviation could potentially be reduced with additional training. Additionally, Demby et al. \cite{b21} assessed the performance of artificial neural networks for solving the IK problems of robots with 4, 5, 6, and 7 degrees of freedom (DOF). They used mean square error to evaluate the solutions for the same desired outputs and found that the error decreased as the size of the training set increased. However, they noted that creating an effective training set required a considerable amount of time. The neural network is trained using data generated by the forward kinematics to learn the angular joint values in the configuration space. Specifically, the neural network maps the specified end effector position \((x_d, y_d)^T\) to the corresponding joint configuration \(q = [\theta_1, \theta_2, \theta_3]^T\). A multilayer perceptron (MLP) is utilized to solve the inverse kinematics for the lower limb. The structure of the MLP is depicted in Figure \ref{fig:27}. It consists of a two-layer feed-forward network: a hidden layer with 10 interconnected sigmoid neurons and an output layer with 3 linear neurons.
 
\section{Simulation results}
\label{Simulation results}
\begin{figure}
    \centering
    \includegraphics[width=0.7\textwidth]{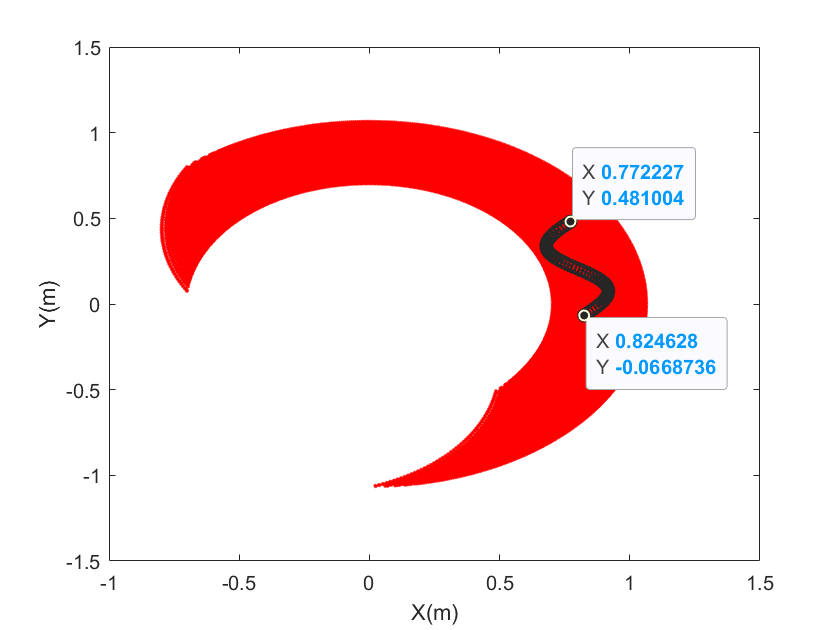}
    \centering
        \caption{Generated trajectory of the lower limb}
    \label{fig:11}
\end{figure}

This section presents a simulation study of human inverse kinematics (IK) using the methods described above. To generate a trajectory, it is essential to ensure that both the starting and final points are within the workspace of the lower limb, as illustrated in Figure \ref{fig:11}. According to \cite{b15}, the average walking speed for adults without mobility issues is between \(1\) and \(1.5\ \text{m/s}\). For this application example, the velocity and acceleration at the beginning and end of the motion are \(1.33\ \text{m/s}\) and \(0\ \text{m/s}^2\), respectively, as shown in Table \ref{tab:3}.
 \\

\begin{figure}
    \centering
    \includegraphics[width=0.7\textwidth]{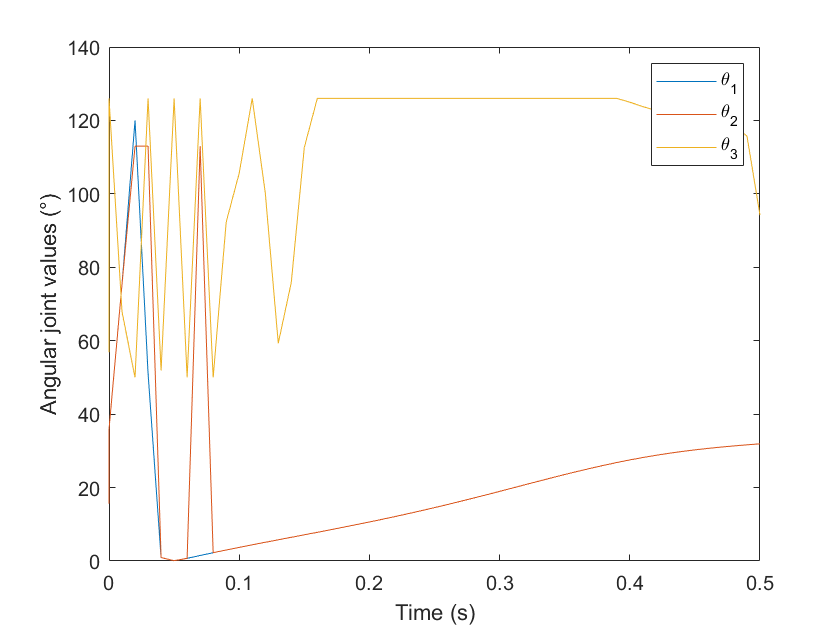}
    \caption{Joints angular tracking using CCD method. For this simulation, $\theta_1^{min}=0^{\circ}$, $\theta_1^{max}=120^{\circ}$, $\theta_2^{min}=0^{\circ}$, $\theta_2^{max}=117^{\circ}$, $\theta_3^{min}=51^{\circ}$ and $\theta_3^{max}=126^{\circ}$.}
    \label{fig:52}
   \end{figure}

\begin{figure}
    \centering
    \includegraphics[width=0.7\textwidth]{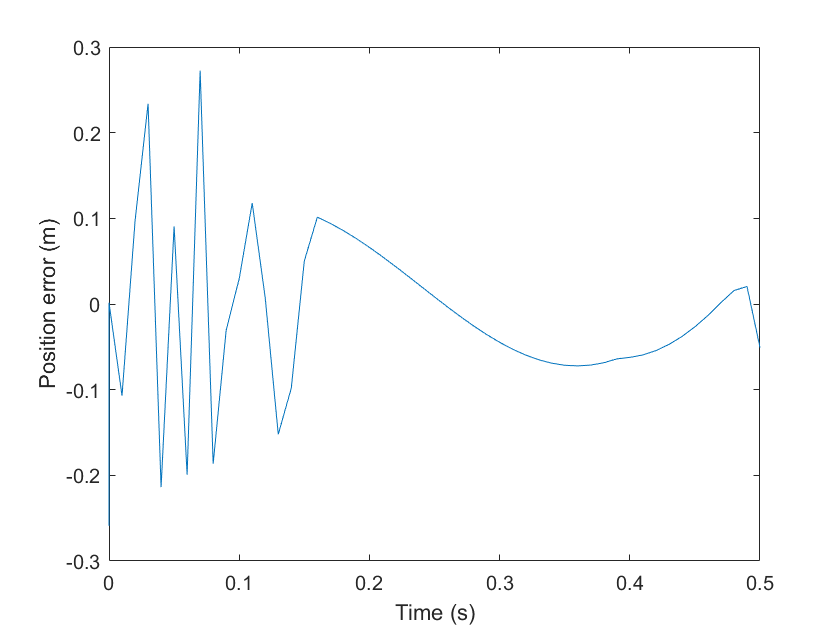}
    \caption{Position error using CCD method}
    \label{fig:53}
   \end{figure}

\begin{figure}
    \centering
    \includegraphics[width=0.7\textwidth]{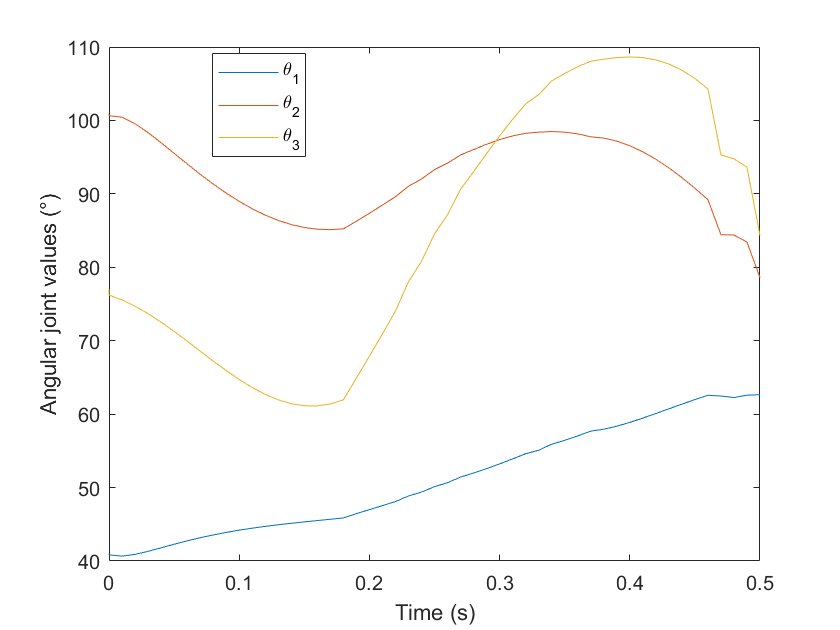}
    \caption{Joints angular tracking using MPPI method. For this simulation, $\theta_1^{min}=42^{\circ}$, $\theta_1^{max}=63^{\circ}$, $\theta_2^{min}=79^{\circ}$, $\theta_2^{max}=102^{\circ}$, $\theta_3^{min}=61^{\circ}$ and $\theta_3^{max}=109^{\circ}$.}
    \label{fig:303}
\end{figure}

\begin{figure}
    \centering
    \includegraphics[width=0.7\textwidth]{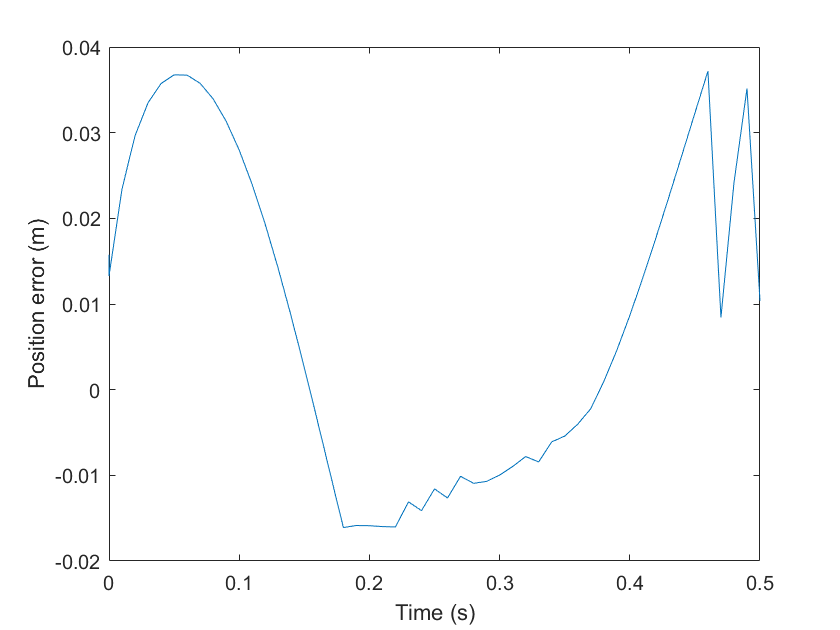}
    \caption{Position error using MPPI method}
    \label{fig:28}
   \end{figure}

\begin{figure}
    \centering
    \includegraphics[width=0.7\textwidth]{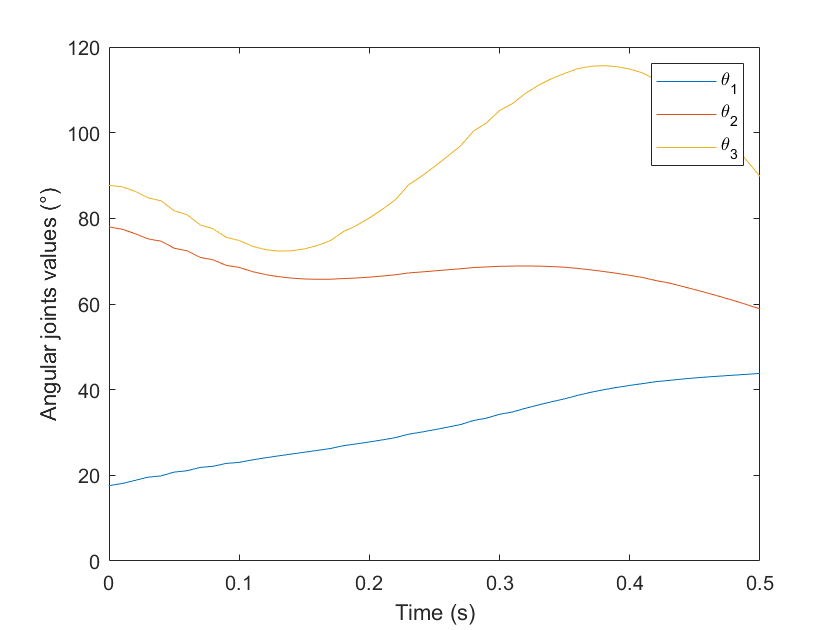}
        \caption{Joints angular tracking using LMDLS method. For this simulation, $\theta_1^{min}=18^{\circ}$, $\theta_1^{max}=43^{\circ}$, $\theta_2^{min}=61^{\circ}$, $\theta_2^{max}=79^{\circ}$, $\theta_3^{min}=87^{\circ}$ and $\theta_3^{max}=118^{\circ}$.}
    \label{fig:18}
\end{figure}

\begin{figure}
    \centering
    \includegraphics[width=0.7\textwidth]{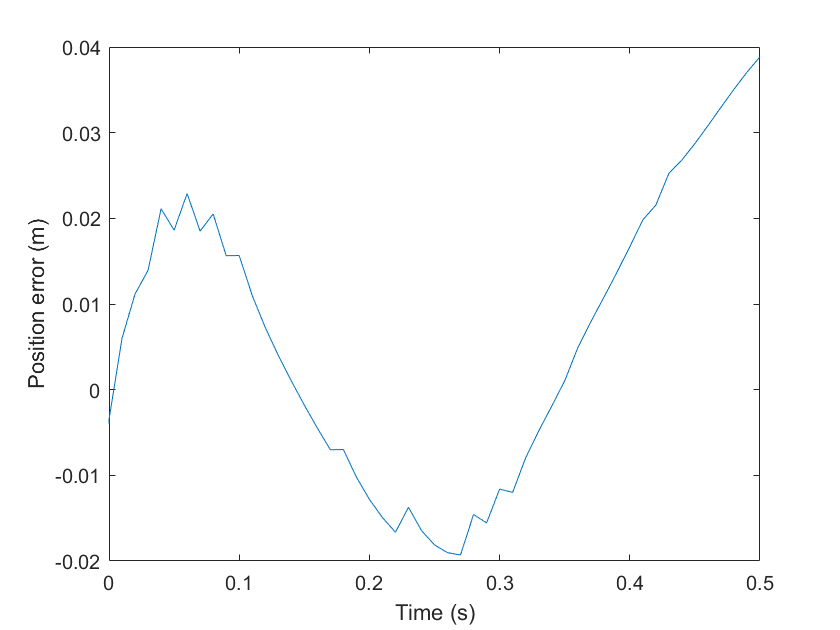}
        \caption{Position error using LMDLS method}
    \label{fig:41}
\end{figure}

\begin{figure}
    \centering
    \includegraphics[width=0.7\textwidth]{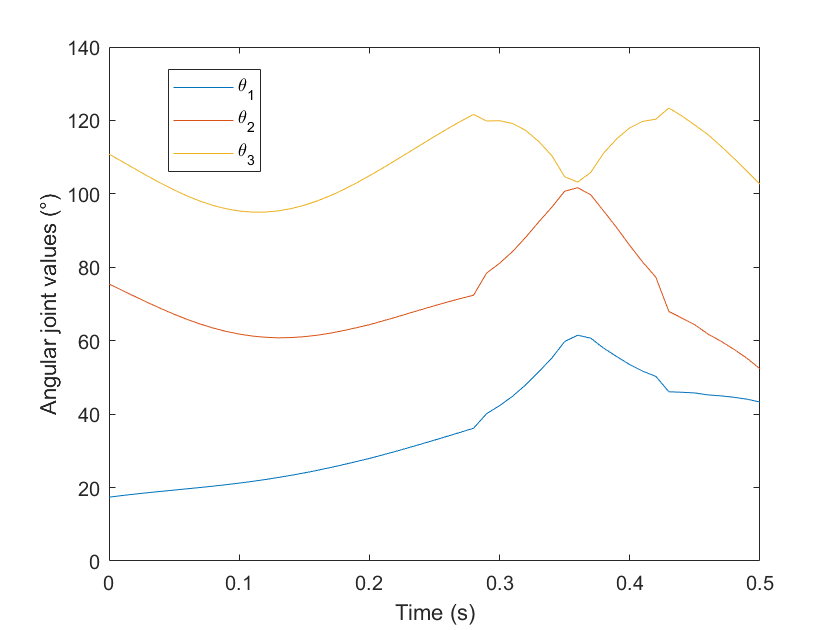}
        \caption{Joints angular tracking using optimisation method. For this simulation, $\theta_1^{min}=18^{\circ}$, $\theta_1^{max}=61^{\circ}$, $\theta_2^{min}=52^{\circ}$, $\theta_2^{max}=101^{\circ}$, $\theta_3^{min}=103^{\circ}$ and $\theta_3^{max}=120^{\circ}$.}
    \label{fig:20}
\end{figure}

\begin{figure}
    \centering
    \includegraphics[width=0.7\textwidth]{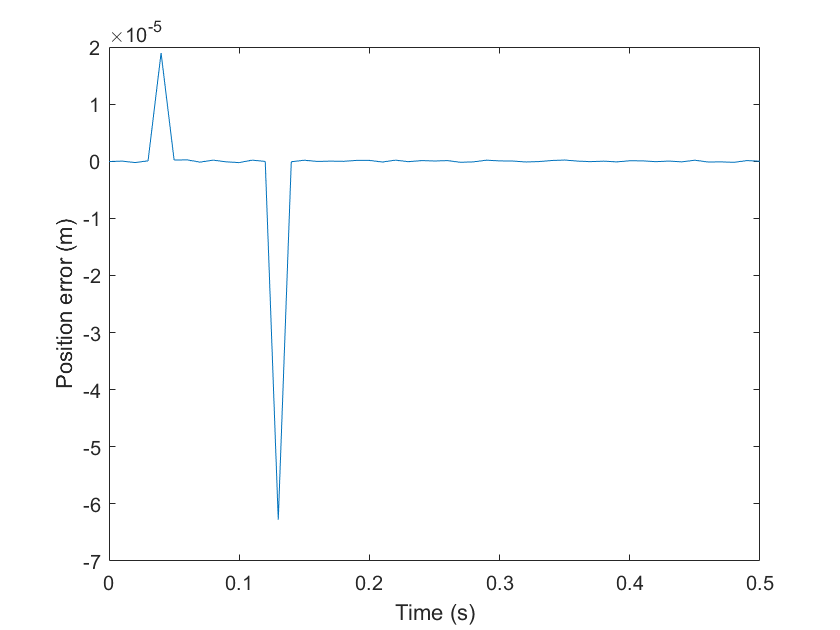}
        \caption{Position error using optimization method}
    \label{fig:19}
\end{figure}

\begin{figure}
    \centering
    \includegraphics[width=0.7\textwidth]{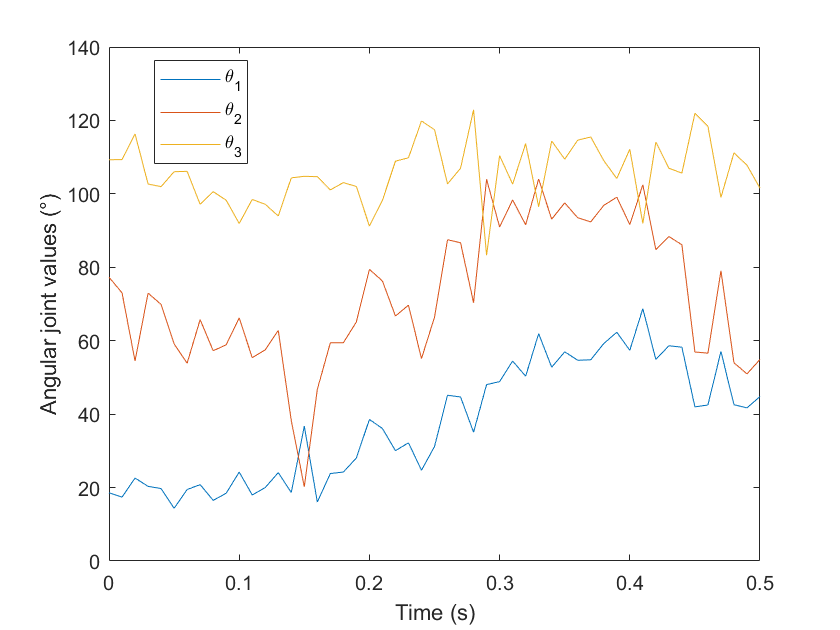}
        \caption{Joints angular tracking using MOOGA method. For this simulation, $\theta_1^{min}=16^{\circ}$, $\theta_1^{max}=68^{\circ}$, $\theta_2^{min}=20^{\circ}$, $\theta_2^{max}=105^{\circ}$, $\theta_3^{min}=84^{\circ}$ and $\theta_3^{max}=120^{\circ}$.}
    \label{fig:21}
\end{figure}

\begin{figure}[htp!]
    \centering
    \includegraphics[width=0.7\textwidth]{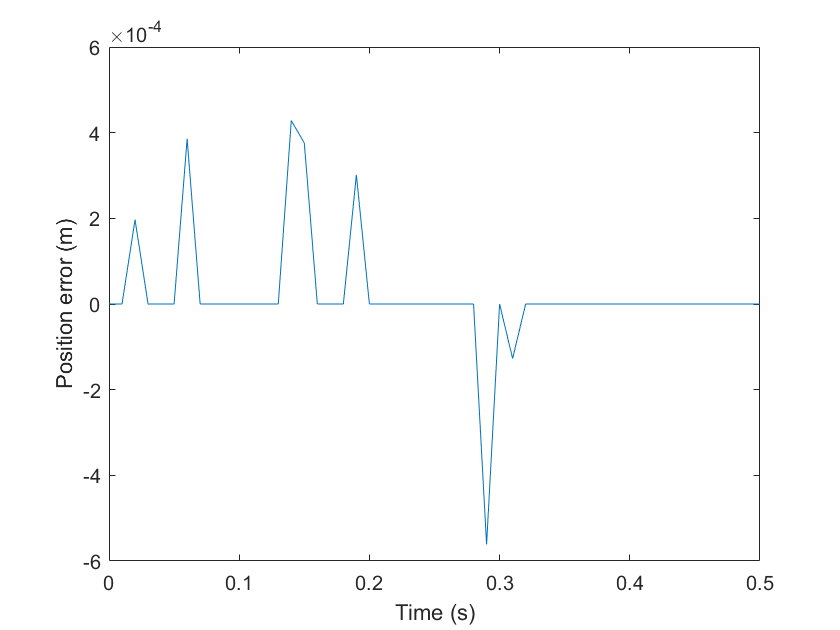}
        \caption{Position error using MOOGA method}
    \label{fig:22}
\end{figure}

\begin{figure}[htp!]
    \centering
    \includegraphics[width=0.7\textwidth]{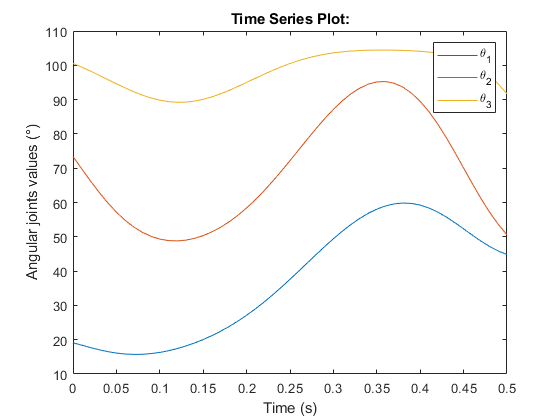}
        \caption{Joints angular tracking using neural network method. For this simulation, $\theta_1^{min}=15^{\circ}$, $\theta_1^{max}=59^{\circ}$, $\theta_2^{min}=48^{\circ}$, $\theta_2^{max}=95^{\circ}$, $\theta_3^{min}=89^{\circ}$ and $\theta_3^{max}=104^{\circ}$.}
    \label{fig:26}
\end{figure}

\begin{figure}[htp!]
    \centering
    \includegraphics[width=0.7\textwidth]{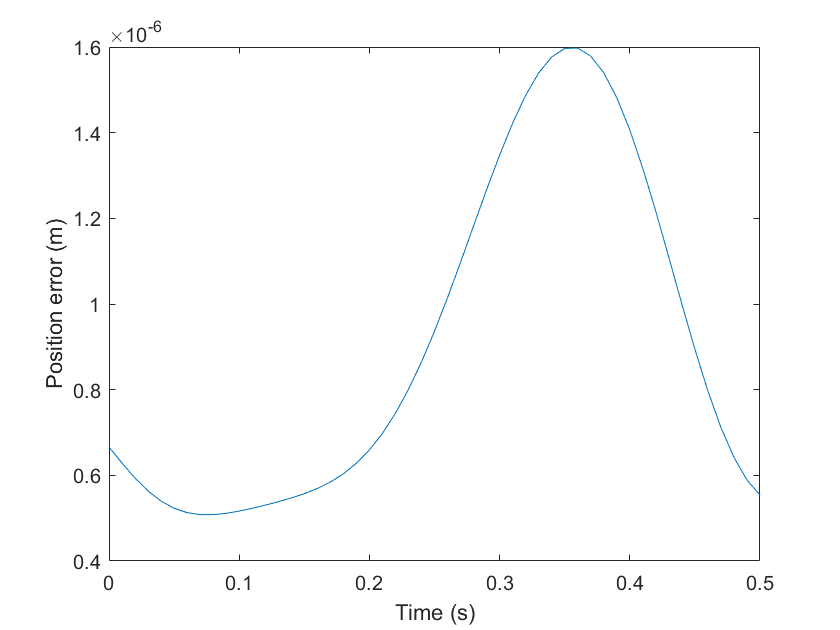}
        \caption{Position error using neural network method}
    \label{fig:25}
\end{figure}

\begin{table}
    \centering
     \caption{Initial and final values for: position, velocity and acceleration of a motion}
    \label{tab:3}
    \begin{tabular}{|cc|c|c|}
      \hline
        & & x & y \\
        \hline
         Position ($m$) & Initial & 0.824628 & -0.0668736   \\
          & Final & 0.772227 & 0.481004   \\
           \hline
         Speed ($m/s$) & Initial & 1.33 & 1.33  \\
           & Final & 1.33 & 1.33  \\
            \hline
         Acceleration ($m/s^2$) & Initial & 0 & 0 \\
          & Final & 0 & 0 \\
         \hline
    \end{tabular}
\end{table}
Thus, assuming that the movement lasts \(0.5\) seconds, the results of this simulation are illustrated in Figure \ref{fig:11}.
The results of the inverse kinematics simulation for the lower limbs using the Cyclic Coordinate Descent (CCD) method are shown in Figure \ref{fig:52}. To determine the position error illustrated in Figure \ref{fig:53}, we used these results along with the forward kinematics described in Equation (\ref{eq:5}) to generate the end effector trajectory. However, the Matlab simulation indicates that this method causes the links to rotate in a sequence that does not match natural movement, resulting in an unnatural appearance of the lower limb's motion. Therefore, the CCD method fails to account for the physiological constraints of the human leg.
In contrast, Figure \ref{fig:303} presents the simulation results for the inverse kinematics of the lower limb using the Moore-Penrose Pseudo-Inverse (MPPI) algorithm. As depicted in Figure \ref{fig:28}, the position error is approximately \(0.04\ \text{mm}\). However, this algorithm does not account for angular constraints.
\\
Regarding the Levenberg-Marquardt Damped Least Squares (LMDLS) technique, the simulation results are shown in Figures \ref{fig:18} and \ref{fig:41}. This method incurs a high computational cost, primarily due to the complexity of the human leg's structure. Figure \ref{fig:20} displays the angular joint values obtained using the optimization algorithm. Figure \ref{fig:19} indicates that the position error is nearly zero, demonstrating the high accuracy of the optimization technique. Figures \ref{fig:21} and \ref{fig:22} illustrate the joint trajectory and position error obtained using the Multi-Objective Optimization Genetic Algorithm (MOOGA) technique, with the same objective function and constraints as described in subsection \ref{Optimization method}. Although MOOGA is accurate, it exhibits abrupt and significant variations in angular configurations. For the neural network approach, out of 127,282 samples collected, 15\% were used for validation, 15\% for testing, and 70\% for training the neural networks. The results for joint angular values and position error are presented in Figures \ref{fig:26} and \ref{fig:25}, respectively. As shown in Figure \ref{fig:25}, the neural network technique achieves high accuracy with an error of less than \(1.6 \times 10^{-6}\).


\section{Discussion}
\label{Discussion and Conclusion}
It is inferred from TABLE \ref{tab:5}, that neural networks method root mean square position error is less compared to other methods, but optimization and  MOOGA methods also show accurate results where root mean square position error is less than $2.10^{-4}$. Moreover, CCD and neural networks methods are fast computationally compared to other methods. 
\begin{equation}
    I_{c}= mean\left(\underbrace{\xi\sum_{i=1}^{3}\bigg\vert\frac{d^3\theta_i(t)}{dt^3}\bigg\vert}_{Energy} +\underbrace{ \mu D_{CoM}+\beta\sum_{i=1}^{3}-\left(\log(-(\theta_i(t)-\theta_i^{max}))+\log(-(\theta_i^{min}-\theta_i(t)))\right)}_{Robustness}\right)
    \label{eq:55} 
\end{equation}
Thereafter, the comfort index $I_c$ shown in Equation (\ref{eq:55}) is used to evaluate the body posture, and it consists of two components:

\begin{itemize}
    \item Energy : lower limb must satisfy during walking the minimum energy constraint which is related to the minimum jerk approach.
      \item Robustness : the lower limb posture is represented by the generalized coordinates $q=\left[\begin{array}{ccc} \theta_1 & \theta_2 & \theta_3 \end{array}\right]^T $, which is constrained by upper and lower limits. Hence, the logarithmic function which is given by Equation (\ref{log}), increases remarkably when joint angles approach their respective barriers. Besides, the distance between the center of mass of the lower limb and that of whole body given by Equation (\ref{eq:06}), must be minimal in order to overcome the fatigue and the musculoskeletal discomfort. Let $M_i^{seg}$ and $CoM_i$ represent the mass and the coordinate of the center of mass of segment i, respectively, with $i\in [1,2,3]$, then:
\end{itemize}
\begin{equation}
    D_{CoM}= \left\vert \frac  {\sum_{i=1}^{3}M_i^{seg} CoM_i}{\sum_{i=1}^{3} M_i^{seg}} \right\vert
    \label{eq:06}
    \end{equation}
With $\xi$, $\mu$ and $\beta$ are homogenization of the comfort index coefficients.
The results in Tab. \ref{tab:5} show a significantly lower comfort index for neural networks method. For cyclic coordinate descent method, the comfort index goes to infinity because the joint angles are near their limits during the motion as depicted in Figure \ref{fig:52}.
\begin{figure}[htp!]
    \centering
    \includegraphics[width=0.7\textwidth]{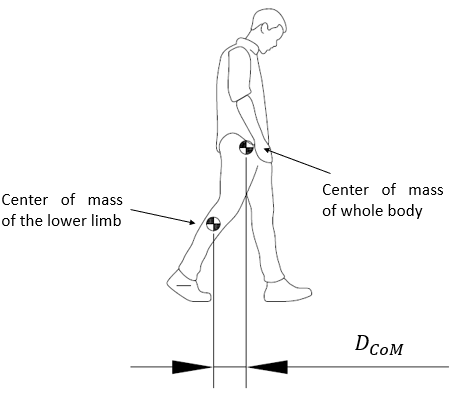}
    \caption{ Positions of the centers of mass in human body }
    \label{fig:my_label}
\end{figure}
\begin{table}
    \centering
    \caption{Comparative table of IK methods}
    \begin{tabular}{|c|c|c|c|}
      \hline
        IK method & Computational time (s) & RMSE & $I_c$\\
         \hline
        CCD method & $0.006101$ & $0.0982$ & - \\
        \hline
        MPPI method & $2.777852 $ & $0.0603$ & 1.1530\\
          \hline
         LMDLS method & $2.051712$ & $0.0299$ & 1.1551\\
         \hline
         Optimization method & $0.0363921$ & $9.0951.10^{-6}$ & 1.1514\\
           \hline
         MOOGA method & $0.5939144$ & $1.3407.10^{-4}$ & 1.1577\\
         \hline
         NN method & $0.0082538$ & $9.7244.10^{-7}$ &  1.0256\\
         \hline
    \end{tabular}
    \label{tab:5}
\end{table}
\section{Conclusion}
This paper presents a comparative study of inverse kinematics techniques for human lower limbs. Theoretical results indicate that the neural network method outperforms the other methods in terms of root mean square position error, computational time, and the generation of realistic postures. The comfort of human motion is influenced by physiological factors such as energy consumption, joint angle limits, and the distance between the center of mass of the lower limb and the entire body. Additionally, comfort posture is affected by environmental and psychological factors, which are not addressed in this study.

\bibliographystyle{unsrt}  
\bibliography{Manuscript}  


\end{document}